\documentclass[conference,10pt]{IEEEtran}
\ifCLASSINFOpdf
    \usepackage[pdftex]{graphicx}
\else
    \usepackage[dvips]{graphicx}
\fi
\usepackage{caption}
\usepackage{subcaption}

\usepackage{amsmath}
\usepackage{amssymb}
\usepackage{hyperref}

\usepackage[ruled,vlined,linesnumbered]{algorithm2e}

\usepackage{xcolor}
\usepackage{multirow}

\DeclareMathOperator*{\argmin}{arg\,min}
\usepackage{caption}

\IEEEoverridecommandlockouts
\IEEEpubid{\makebox[\columnwidth]{ 979-8-3503-1175-4/23/\$31.00 $ \copyright$2023 IEEE 
\hfill{ \hspace{\columnsep}\makebox[\columnwidth] {}}
}} 
\begin{document}

%

\title{CARMA: Context-Aware Runtime Reconfiguration for Energy-Efficient Sensor Fusion}






\author{\IEEEauthorblockN{Yifan Zhang\textsuperscript{\textsection}, Arnav Vaibhav Malawade\textsuperscript{\textsection}, Xiaofang Zhang, Yuhui Li, DongHwan Seong, \\Mohammad Abdullah Al Faruque, Sitao Huang}
\IEEEauthorblockA{
Department of Electrical Engineering and Computer Science, University of California, Irvine \\
\{yifanz58, malawada, xiaofaz7, yuhuil10, dseong1, alfaruqu, sitaoh\}@uci.edu}\vspace{-6mm}}



%

\maketitle
\begingroup\renewcommand\thefootnote{\textsection}
\footnotetext{Equal contribution}
\endgroup


\begin{abstract} 
 Autonomous systems (AS) are systems that can adapt and change their behaviors in response to unanticipated events and include systems such as aerial drones, autonomous vehicles, and ground/aquatic robots. AS require a wide array of sensors, deep learning models, and powerful hardware platforms to perceive the environment and safely operate in real-time. 
However, in many contexts, some sensing modalities negatively impact perception while increasing the system's overall energy consumption. 
Since AS are often energy-constrained edge devices, energy-efficient sensor fusion methods have been proposed.
However, existing methods either fail to adapt to changing scenario conditions or to optimize system-wide energy efficiency.
We propose CARMA, a context-aware sensor fusion approach that uses context to dynamically reconfigure the computation flow on a field-programmable gate array (FPGA) at runtime.
By clock gating unused sensors and model sub-components, CARMA significantly reduces the energy used by a multi-sensory object detector without compromising performance.
We use a deep learning processor unit (DPU) based reconfiguration approach to minimize the latency of model reconfiguration. 
We evaluate multiple context identification strategies, propose a novel system-wide energy-performance joint optimization, and evaluate scenario-specific perception performance.
Across challenging real-world sensing contexts, CARMA outperforms state-of-the-art methods with up to 1.3$\times$ speedup and 73\% lower energy consumption.
\end{abstract} 


\section{Introduction}
Autonomous systems (AS) 
radically improve productivity, logistics, and safety by enabling systems such as aerial drones, ground and aquatic robots, and consumer autonomous vehicles (AVs) to operate without direct human control. 
These applications require closely coupled perception and state estimation algorithms to navigate complex and unpredictable real-world scenarios in real time.
Advanced deep learning models and multiple heterogeneous sensors (cameras, radars, and LiDARs) are necessary for perception across different weather and lighting conditions. 
However, the increasing complexity of modern AS comes with rising energy costs \cite{lin2018architectural}, which can be fatal for energy-constrained AS. The thermal design power of modern AS System-on-Chips (SoCs) 
can exceed 800 W~\cite{Abuelsamid2020orin}, and the combined sensing, computation, and thermal loads can reduce operating range by over 11.5\% \cite{ he2022energy}. 

Since the perception system is a major energy consumer in AS \cite{lin2018architectural, malawade2021sage}, several efficient sensor fusion methods have been proposed. However, these methods use static architectures (\textit{e.g.}, early or late fusion) that can fail in complex visual contexts where one or more sensors may be compromised \cite{malawade2022hydrafusion}. To address these limitations, context-aware dynamic architectures for sensor fusion have been proposed \cite{malawade2022hydrafusion, malawade2022ecofusion}, where the model adapts to changing environmental conditions to enable robust and energy-efficient perception across diverse sensing conditions.
Still, existing methods only focus on reducing algorithmic energy usage and ignore large energy consumers, such as the sensors and the hardware computation platforms. 

In summary, the key challenges include:
(i) effectively perceiving complex and adverse driving scenarios;
(ii) reducing the energy consumption of the complete perception system, including sensors, hardware, and algorithms; and
(iii) adapting the system configuration to different contexts, improving energy efficiency without compromising performance.

To overcome these challenges, we propose CARMA, a context-aware dynamic sensor fusion approach that uses \textit{runtime model reconfiguration} to adapt its architecture on an FPGA. CARMA uses deep learning processing unit (DPU)~\cite{dpudatasheet} on FPGA for efficient, low-latency runtime reconfiguration. 
CARMA implements a tunable energy-performance optimization over the whole system, including sensors, model architecture, and hardware platform, to maximize energy savings without compromising performance. 
To our knowledge, this is the first work to propose energy-efficient sensor fusion via context-aware runtime model reconfiguration on FPGAs.

Our major contributions can be summarized as follows:
\begin{enumerate}
    \item We propose CARMA, an approach for dynamically reconfiguring a complete sensor fusion system for object detection at runtime using contextual information. CARMA uses DPUs on FPGA to enable runtime model reconfiguration with negligible model switching latency. 
    \item We propose a method for intermittently performing context identification to enable intelligent sensor and submodel clock gating to maximize energy efficiency.
    \item We use tunable joint optimization between perception performance and energy consumption to maximize energy efficiency while minimizing perception impacts. 
    \item We show that CARMA significantly reduces system-wide energy usage compared to state-of-the-art sensor fusion methods and achieves equivalent or better object detection performance across diverse autonomous driving scenarios with up to $\mathbf{1.3 \times}$ inference speedup and 73\% lower energy consumption.
\end{enumerate}


\section{Related Works}

\subsection{Adaptive Computing Systems on FPGA}
\label{subsec:Adaptive_Systems_FPGA}

Self-adaptive systems can modify their runtime behavior according to changing environments and system goals. 
\cite{irmak2021increasing} presents a dynamically reconfigurable convolutional neural network (CNN) accelerator optimized for throughput. In \cite{youssef2020energy}, an FPGA reconfigures at runtime to use a lower power design when the battery level decreases. 
However, its reconfiguration latency is proportional to the bitstream size, which limits it from applying to large components. The DPUs enable users to reconfigure CNN models at runtime with minimal latency overhead. \cite{electronics11010105} explored a DPU-based energy-efficient hardware accelerator. However, it does not optimize energy efficiency system-wide or handle complex environments.


\subsection{Energy-Performance Optimization}
\label{subsec:rw-energy-perf-optim}
Several works have explored methods on energy-performance trade-off of deep learning algorithms at runtime targeting single-modality  image classification task \cite{mullapudi2018hydranets,takhirov2016energy,hao2019codesign}. Recent works have extended these optimizations to multi-sensor fusion for perception \cite{balemans2020resource}. \cite{malawade2022ecofusion} proposes a dynamic-width sensor fusion model that aims to select lower energy submodels while maintaining performance. Although this approach incorporates multimodality, it only optimizes the object detection model parameters and omits \textit{system-wide} energy optimizations. In contrast, we propose using runtime model reconfiguration on a heterogeneous FPGA-driven computing platform to maximize the energy saved by dynamic model selection while applying system-wide energy optimizations to reduce energy usage further.

\subsection{Intermittent Sensing and Control in Autonomous Systems}
\label{subsec:intermittent}
Due to the energy constraints of many AS, several methods for intermittent sensing and control have been proposed to reduce energy consumption without compromising performance~\cite{gokhale2021feel, huang2020opportunistic}. 
\cite{dash2021intermittent} proposes using an intermittent control strategy for autonomous driving to emulate human-like control behavior. 
Like these works, CARMA targets energy efficiency by intermittently reconfiguring the model architecture and the set of active sensors to match the current environment context.


\section{Methodology}

\subsection{Problem Formulation}

\subsubsection{Object Detection Model}
AS use object detection to avoid collisions, predict motion, and enable safe path planning. The goal of the object detector $\phi$ is to use the sensor measurements $\textbf{X}$ to accurately identify the objects $\textbf{Y}$ in the environment:
\vspace{-2mm} 
\begin{equation}\vspace{-1ex}
    \mathbf{Y} = \phi(\mathbf{X}), \; \text{where} \;
    \mathbf{Y} = \{\mathbf{Y}_{class}^i,  \mathbf{Y}_{reg}^i\}_{i=1 \dots d}
\end{equation}
where $\mathbf{Y}_{class}^i,  \mathbf{Y}_{reg}^i$ denote the class and bounding box, respectively, of object $i$.
Extending this framework to multi-sensor perception, early fusion across $s$ sensors can be modeled as: 
\vspace{-2mm} 
\begin{equation}\vspace{-1ex} 
    \mathbf{\hat{Y}}  = \phi(\psi(\mathbf{X}_1, \mathbf{X}_2, \dots , \mathbf{X}_s )),
\end{equation}

where $\psi$ is the function for fusing the sensor data before the object detector processes it, and $\hat{Y}$ represents the object predictions. 
Similarly, late fusion across $s$ sensors can be modeled as fusing the outputs of an ensemble of independent object detectors:
\vspace{-2mm} 
\begin{equation}
    \mathbf{\hat{Y}}_1,\, \mathbf{\hat{Y}}_2,\, \dots, \mathbf{\hat{Y}}_s = \phi_1(\mathbf{X}_1),\,  \phi_2(\mathbf{X}_2),\, \dots , \, \phi_s(\mathbf{X}_s)
\end{equation}
\vspace{-6mm}
\begin{equation}\vspace{-1ex}
    \mathbf{\hat{Y}} = \phi_{f}(\mathbf{\hat{Y}}_1, \mathbf{\hat{Y}}_2, \dots , \mathbf{\hat{Y}}_s),
\end{equation}
where $(\phi1, \phi_2, ..., \phi_s)$ represent independent object detectors, and $\phi_{f}$ represents the late fusion function for combining their outputs.
Our proposed approach uses context to identify the best combination of early and late fusion to improve the accuracy of the resultant predictions across driving contexts.
As such, the object detection model becomes:
\vspace{-2mm}
\begin{equation}\vspace{-2mm}
    \mathbf{\hat{Y}} = \phi_{f}(\phi_1(\mathbf{X}_1),\,  \phi_2(\mathbf{X}_2),\, \dots, \, \phi_3(\psi(\mathbf{X}_2,\, \mathbf{X}_s)))
\end{equation}
Where $\phi_1$ and $\phi_2$ represent single-sensor object detectors, $\phi_3$ is a multi-sensor object detector using early fusion, and $\phi$ is the late fusion function for fusing the detectors' outputs to obtain $\hat{Y}$. Section \ref{subsec:gating} describes how CARMA identifies context and selects the appropriate model configuration.

\subsubsection{Energy Model}
We model the energy usage of the complete AV driving system $E_{sys}$ as the total energy consumed by the sensors $E_s$ and the execution of the algorithm $E_a$ on the hardware platform. 
\vspace{-2mm} 
\begin{equation}\vspace{-1ex}
    E_{sys} = E_s + E_a
\end{equation}
We omit factors such as drivetrain energy usage and battery lifetime as these factors have been studied in existing work \cite{vatanparvar2018extended, baek2018battery} and can be used in conjunction with our approach.
Typical AS contain some combination of static sensors (\textit{e.g.}, cameras, ultrasonic sensors, front-facing radar) and rotating sensors (\textit{e.g.}, spinning top-mounted LiDAR). The energy consumption per sensor $s \in S$ can be computed from the measurement power $P_s^{meas.}$, measurement frequency $f_s$, and, for spinning sensors, the motor power $P_s^{motor}$, as follows:
\vspace{-2mm}
\begin{equation}\vspace{-1ex}
    E_s = (P_s^{meas.} + P_s^{motor}) * 1/f_s 
\end{equation}
To reduce the energy consumption of the complete system, we clock gate sensors unused in the current visual context. The LiDAR and radar sensors in our testbed, discussed in Section \ref{subsec:exp_setup}, are top-mounted spinning sensors, while the cameras are fixed sensors without motors. Since the LiDAR and radar have inertia and require several seconds to start and stop rotating, we assume that we only clock gate the measurement components while keeping the motor spinning so they can be quickly re-enabled to ensure safety. 
The total power consumption of the Navtech CTS350-X radar is 24~W \cite{navtechdatasheet}, while the Velodyne HDL-32E LiDAR uses 12~W \cite{velodynedatasheet} and the ZED camera uses 1.9~W \cite{zeddatasheet}.  The Navtech CTS350-X needs 2.4~W to spin the motor, so  $P^{meas.}_{radar} = 21.6$~W. Using comparable LiDAR motor models for the Velodyne HDL-32E, we estimate $P^{meas.}_{LiDAR}=9.6$~W. 

Since our object detection model is reconfigurable, the algorithm energy consumption $E_a$ can be computed as:
\begin{equation}
    E_a(\phi, \mathbf{X}) = P_a(\phi, \mathbf{X}) * t(\phi, \mathbf{X}),
\end{equation}
where $t(\phi, \mathbf{X})$ represents the processing latency in seconds and $P_a(\phi, \mathbf{X})$ represents the power consumption in Watts of processing input $\mathbf{X}$ through the current model configuration $\phi$ on the hardware platform. We measured the power and latency of each model configuration on our hardware platform, the Xilinx Kria KV260 FPGA, to compute $E_a$ offline for use in our multi-objective optimization.

\subsubsection{Multi-Objective Optimization}
We implement a tunable joint optimization between system-wide energy consumption and model performance to enable our approach to minimize energy without compromising performance. Since there is typically a trade-off between these two objectives, we use a $\lambda_E$ term to allow model designers to specify the preference for energy efficiency over performance depending on the application of the system. Given that we know the expected prediction performance $L$ of configuration $\phi$ for an input $\textbf{X}$, denoted as $L(\phi, \textbf{X})$, and the expected system-wide energy consumption of that configuration $E_{sys}(\phi, \textbf{X})$, our optimization can be formulated as:
\begin{equation}
    L_{opt}(\phi, \textbf{X}) =  L(\phi, \textbf{X}) * (1-\lambda_E) + E_{sys}(\phi, \textbf{X}) * \lambda_E
\end{equation}
\vspace{-6mm}
\begin{equation}
    \phi^*(\textbf{X}) = \argmin_{\phi \in \Phi}(L_{opt}(\phi, \textbf{X})), 
\end{equation}
where $\phi^*(\textbf{X})$ represents the model configuration that best minimizes the joint optimization loss $L_{opt}$ for input $\textbf{X}$ for the given $\lambda_E$. \cite{malawade2022ecofusion} used a similar optimization to select which branches to execute, with all other system components remaining fixed. However, our proposed approach includes clock gating of unused sensors and stems, drastically increasing the potential energy savings and enabling system-wide optimization.

\subsection{System Architecture}
\label{subsec:system_architecture}
\begin{figure}
    \centering
    \includegraphics[width=\linewidth]{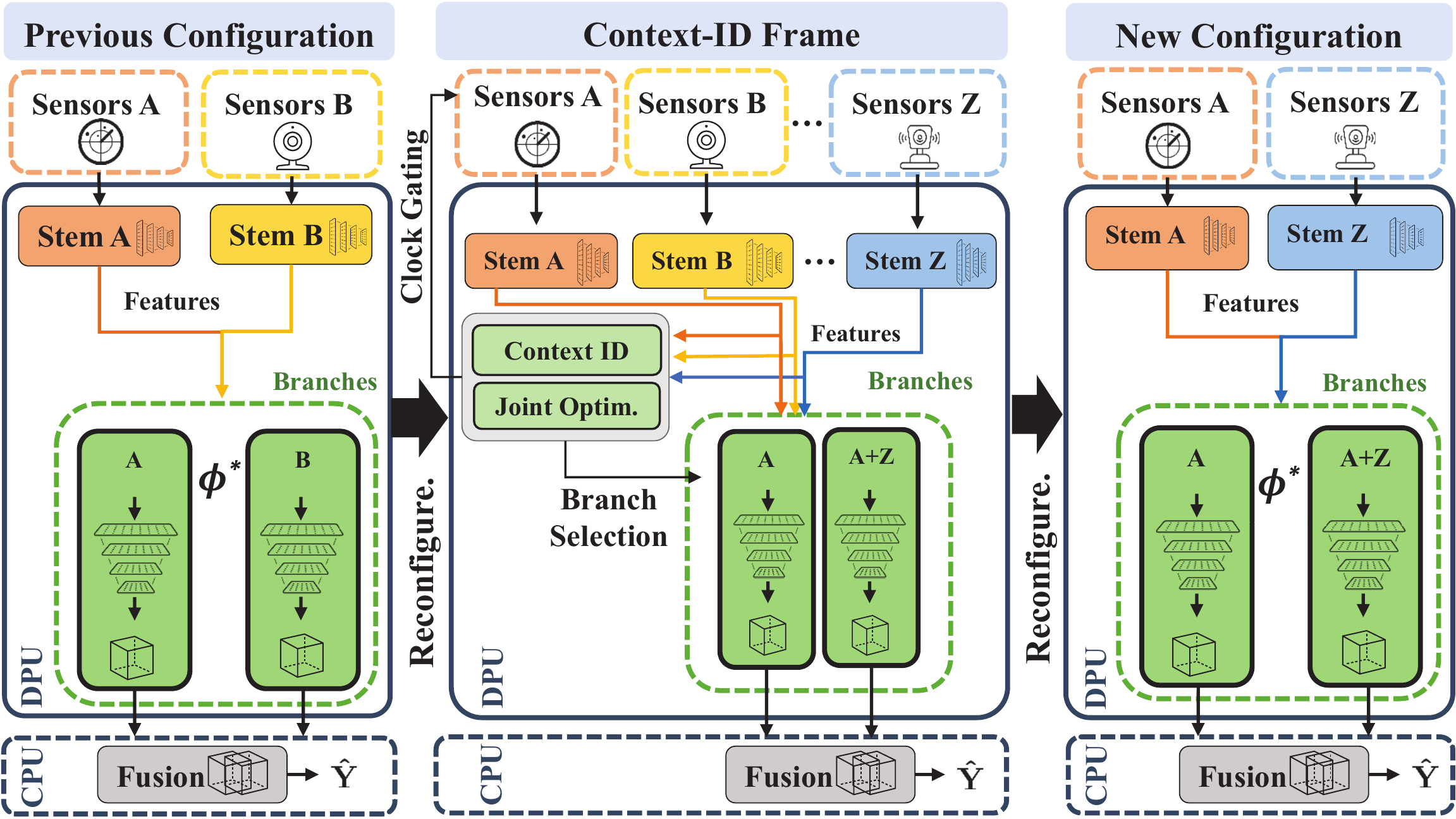}
    \caption{CARMA System Architecture and Reconfiguration Workflow}
    \label{fig:archi}
    \vspace{-4mm}
\end{figure}

CARMA's architecture is shown in Fig. \ref{fig:archi}. CARMA consists of a runtime reconfigurable multi-branch sensor fusion model for object detection. Section \ref{subsec:hardware_execution_model} elaborates on our runtime reconfiguration approach on hardware, while the following text describes our sensor fusion model.
The model consists of four key components, (i) feature extraction, (ii) context identification, (iii) submodel selection, and (iv) output fusion. First, multi-modal sensor data is processed by modality-specific \textit{Stem} models to extract an initial set of features for each sensor. These features are then used by the \textit{Gate} model to identify the current visual context. This context is used to select the set of submodels (\textit{Branches}) to execute that optimizes performance and energy efficiency. Each active branch outputs a set of object detections collected and fused by the \textit{Fusion Block} to produce a final set of refined detections. 

\subsubsection{Stem and Branches}
\label{subsec:stem_branches}

We utilize the single shot multibox detector (SSD) \cite{liu2016ssd} for object detection, known for its superior speed and performance compared to Faster R-CNN \cite{ren2015faster}. SSD employs a single-pass CNN to perform region proposal and object detection, eliminating the need for a separate Region Proposal Network. With a smaller model size and fewer intermediate feature maps, SSD requires fewer hardware resources and has lower memory bandwidth, making it faster to execute on FPGAs. Our proposed architecture incorporates SSD's ResNet-18 backbone, using the first six layers as modality-specific preprocessors (\textit{stem}) and the remaining 23 layers as branches. We implement single-sensor branches for four inputs (two cameras, one LiDAR, and one radar) and three early-fusion branches that take multiple sensors as input: dual camera, LiDAR and radar, and dual camera with LiDAR. These branches include a single merge convolution layer to combine the sensors across the channel dimension before continuing with processing.



\subsubsection{Context Identification and Gating}
\label{subsec:gating}
To identify the current visual context and perform branch selection, we propose three variants of context-identification, or \textit{gate}, models. The \textit{knowledge} gate uses fixed domain-knowledge rules to select submodels using external contextual information (e.g., weather, time of day, road type). The rules encode domain knowledge on the sensor modalities least likely to be degraded by current environmental factors such as rain, snow or fog.
The \textit{deep} gate uses a 3-layer CNN to infer the current context from the stem output features and directly output the set of branches it infers will perform best in the current visual context. Here, context refers to an abstract visual state estimate generated within the CNN's hidden layers, while the gate output indicates which branches to execute. The \textit{attention} gate is the same as the \textit{deep} gate with the addition of a self-attention layer.
Given the set of all possible model configurations $\Phi$, the objective of the gate is to estimate the performance $L$ of each configuration $\phi$ for the current set of input features $F$:
\vspace{-1ex}
\begin{equation}
L(\Phi, \textbf{F}) = \pi(\phi, \textbf{F}),  \forall \phi \in \Phi
\end{equation}
\vspace{-6mm}
\begin{equation}
    \rho(L(\Phi, \textbf{F}), \gamma) = \left\{\phi \in \Phi \; \text{s.t.} \; L(\phi, \textbf{F}) \leq L(\phi', \textbf{F}) + \gamma\right\}
\end{equation}
\vspace{-6mm}
\begin{equation}
\label{eqn:rho}
    \Phi^* = \rho(L(\Phi, \textbf{F}), \gamma), 
\end{equation}
where $\pi$ represents the gating model and $\rho$ represents a function for identifying the set $\Phi^*$ of top performing configurations with an estimated error within $\gamma$ of the best configuration $\phi'$.

\subsubsection{Fusion Block}

The fusion block in CARMA combines object detections from active branches to produce more accurate final bounding box predictions. We employ \textit{weighted boxes fusion} \cite{solovyev2021weighted}, which averages proposed boxes based on confidence scores. In CARMA, the fusion block runs on the CPU due to its complex program logic, which is better supported on the CPU than the DPU. It can also utilize idle CPU resources during DPU inference.

\subsection{Hardware Design Choices}
\label{subsec:Hardware design choices}
CARMA is adaptive to various platforms. Still, safety-critical real-time tasks require careful hardware design choices.

\subsubsection{High Throughput}

In autonomous systems (AS), real-time data processing with low latency is crucial for safe and efficient vehicle operation. A minimum rate of 10 frames per second (FPS) \cite{lin2018architectural} is typically required to enable accurate control in dynamic environments. The DPU offers user-configurable parameters for optimizing performance, including pixel, input channel, and output channel parallelism. For different branch configurations, the computation workload can vary from 9 to 27 GOP ($10^{9}$ operations). With tailored parallelism settings of 8, 16, and 16, respectively, the DPU achieves a theoretical speed of 1228.8 GOP/s at a clock frequency of 300 MHz (2$\times$8$\times$16$\times$16 = 4096 operations per cycle), to maintain safe FPS and cover possible tail latency. Additionally, our on-board profiling results show that with 2000 MB/s memory bandwidth, we can guarantee 700 GOP/s average throughput when memory-bounded. (For a 12 GOP workload model with 33 MB estimated memory access).

\subsubsection{Fast Context Switch Interval}
CARMA changes branch configurations (\textit{context switch}) at runtime. Fast context switch intervals are necessary to handle various tasks and events that may occur during vehicle operation. 
CARMA uses Vitis AI Runtime to load the instruction files into the DPU for inference and switch the context by changing the calling threads corresponding to different configurations. Loading of model instruction files and inference are performed simultaneously, reducing the context switch time to the time of the thread switch (less than 1ms), while traditional FPGA runtime reconfiguration waits until the new bitstream is fully deployed on-board. Since each model file is $<$25 MB and our system has 4 GB on-board memory, our system can store all model configurations in DDR memory.

\subsection{Hardware Execution Model}
\label{subsec:hardware_execution_model}

\begin{figure}[!t]
    \centering
    \includegraphics[width=0.85\linewidth]{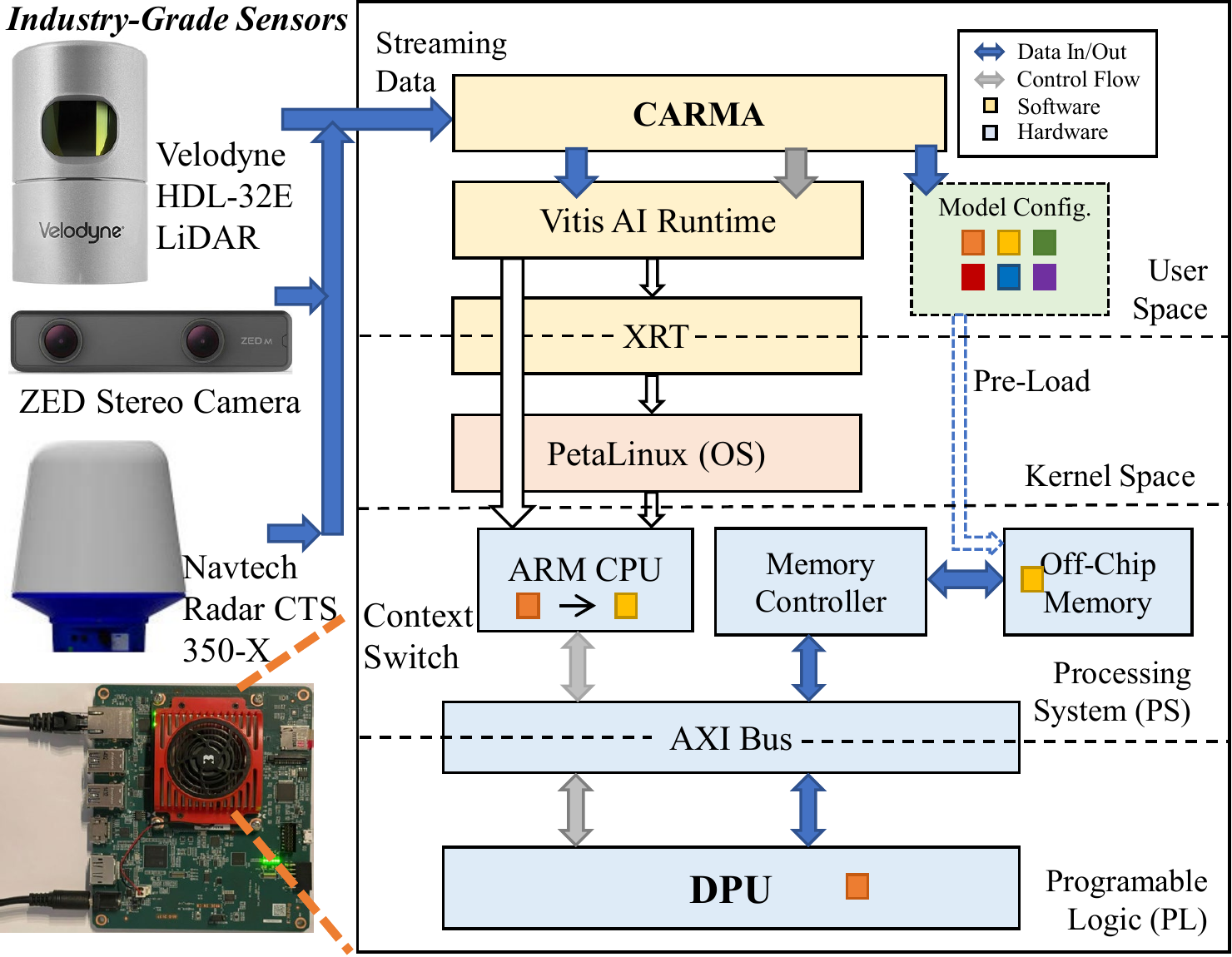}
    \caption{CARMA System Stack and Experimental Testbed} 
    \label{fig:hw_sw_stack}
    \vspace{-2em}
\end{figure}


Fig. \ref{fig:hw_sw_stack} illustrates our hardware execution model. 
CARMA runs in the application layer on PetaLinux and controls our complete sensor fusion system. It uses Vitis AI Runtime, a set of high-level APIs, to interact with the DPU. Xilinx Runtime (XRT) provides a set of low-level APIs that connect the User Space and Kernel Space and control the hardware. The CPU serves as the hardware host control node and controls the DPU, services interrupts, and coordinates data transfers. The processing system (PS) connects to the DPU via the Advanced eXtensible Interface (AXI) bus for transferring data and control signals. When initializing the system, the compiled models for all sensor-fusion configurations are loaded into the off-chip memory, waiting to be called. At runtime, the DPU fetches compiled instructions from off-chip memory to control the operation of its computing engine. 

\subsection{Runtime Workflow and Intermittent Context Identification}
\label{subsec:runtime_methodology}
Several works have demonstrated safe and effective intermittent perception and control approaches, as discussed in Section \ref{subsec:intermittent}.
These approaches are intuitive since real-world visual contexts often remain the same for several seconds, especially in the case of broad visual contexts like \textit{rainy weather} or \textit{night driving}.
We propose using \textit{intermittent} context-identification to enable broader energy optimizations such as clock-gating unused sensors and stem models for brief periods before re-enabling them to identify the current context. 
CARMA can directly integrate with existing methods for safe intermittent perception since they use similar strategies, such as clock gating, to control sensing frequency.

To reduce the overhead of context identification and switching, we propose the Context-ID Frame design, shown in Fig. \ref{fig:archi}.
In sensor fusion mode, we only execute the stems and branches needed for a particular model configuration, minimizing energy consumption. 
In Context-ID mode, we reconfigure the DPU to the Context-ID Frame to select the next model configuration. 
The following two algorithms describe the workflow of our proposed approach. 
\begin{algorithm}
\caption{Runtime Sensor Fusion Algorithm}
\small
\DontPrintSemicolon
\label{alg:sensor_fusion}
\KwIn{$t$, $\phi^*$, $active\_sensors$, $T_c$}
\KwOut{Object Detections $(\mathbf{\hat{Y}})$}

Initialize feature vector $\mathbf{F}$ and branch output vector $\mathbf{\hat{Y}}^*$

\For {s in active\_sensors}{ 
$X_s \gets s(t)$ \tcp*{data input}
$\mathbf{F}[s] \gets stem_s(X_s)$ \tcp*{extract features}}

\For {$\text{branch}$ in $\phi^*$}{
$\mathbf{\hat{Y}}^*[branch] \gets branch(\mathbf{F^*})$\tcp*{pass subset of $\mathbf{F}$}}

$\mathbf{\hat{Y}} \gets fusion\_block(\mathbf{\hat{Y}}^*)$ \tcp*{fuse detections}

\If{$t / T_c = 0$}{
$\phi^*, active\_sensors \gets \mathbf{Algorithm 2}(t+1)$
}
\end{algorithm}
Alg. \ref{alg:sensor_fusion} shows the typical operation of CARMA. For each time step $t$, data is retrieved from the active set of sensors and processed by the current branch configuration $\phi^*$ to produce the output detections $\mathbf{\hat{Y}}$. $T_c$ represents the context re-identification interval; when $t / T_c = 0$, execution transfers to Alg. \ref{alg:context_id} for the next time step $t+1$. 
Here, $T_c$ can be dynamically configured by an intermittent algorithm, such as those from Section \ref{subsec:intermittent}.
In Alg. \ref{alg:context_id}, all sensors and stems are activated, and the sensor features $F$ are passed to the gate module $\pi$ to estimate the loss of each branch configuration. The lowest loss branches are selected by $\rho$ as described in Equation \ref{eqn:rho}. Then, this set $\Phi^*$ is passed to the joint optimization to identify the optimal configuration $\phi^*$. The outputs of the active branches are fused to produce $\mathbf{\hat{Y}}$. 
 After this step, we clock gate the unused sensors, switch to the new model configuration $\phi^*$, and continue executing Alg. \ref{alg:sensor_fusion} with the new $\phi^*$ and $active\_sensors$ at the next time step.

\begin{algorithm}
\caption{Context ID and Reconfigure Algorithm}
\small
\DontPrintSemicolon
\label{alg:context_id}
\KwIn{$t$, $\lambda_E$, $\Phi$, $\gamma$, $E_{sys}(\Phi)$, $all\_sensors$}
\KwOut{Object Detections $(\mathbf{\hat{Y}}), \phi^*, active\_sensors$}

Initialize feature vec. $\mathbf{F}$ and output vec. $\mathbf{\hat{Y}}^*$

\For {s in all\_sensors}{ 
$X_s \gets s(t)$ \tcp*{data input}
$\mathbf{F}[s] \gets stem_s(X_s)$ \tcp*{extract features}}

$L(\Phi) \gets \pi(\mathbf{F},\Phi)$ \tcp*{estimate model losses}
$\Phi^* \gets \rho(L(\Phi), \gamma)$ \tcp*{select candidates}

\For{$\phi$ in $\Phi^*$}{
$L_{joint}(\phi) \gets (1-\lambda_{E}) * L(\phi) + \lambda_{E} * E_{sys}(\phi)$}

$\phi^* \gets \argmin_{\forall \phi \in \Phi^*}(L_{joint}(\phi))$ \tcp*{joint opt.}

$load\_branches(\phi^*)$\tcp*{reconfiguration}

\For {$\text{branch}$ in $\phi^*$}{
$\mathbf{\hat{Y}}^*[branch] \gets branch(\mathbf{F^*})$\tcp*{pass subset of $\mathbf{F}$}
}
$\mathbf{\hat{Y}} \gets fusion\_block(\mathbf{\hat{Y}}^*)$ \tcp*{fuse detections}

Initialize empty set $active\_sensors$

\For{s in all\_sensors}{
\If{$\phi^*$ \text{requires} $s$}{
    $active\_sensors \leftarrow active\_sensors \cup \{s\}$
    }
\Else{
    $clock\_gate(s)$\tcp*{clock gate sensors}
    $disable\_stem(stem_s)$\tcp*{reconfiguration}
    }
}
\end{algorithm}
\vspace{-4mm}
 
\section{Experiments}
\subsection{Experimental Setup}
\label{subsec:exp_setup}


CARMA can be applied to any multi-sensor AS to enable energy-efficient perception. In our experiments, we evaluate CARMA on a popular AS use case: autonomous driving for AVs. 
Our hardware testbed is shown on the left side of Fig. \ref{fig:hw_sw_stack}. 
We use the Xilinx Kria KV260 FPGA as our computing platform. Due to its portability and compatibility, our design could feasibly be implemented on Xilinx automotive-grade FPGAs in a similar manner.
Each model is trained on the RADIATE dataset \cite{sheeny2020radiate}, which contains three hours of high-resolution radar, LiDAR, and stereo camera data across challenging perception contexts. 
We compare against Faster R-CNN object detectors for single sensor inputs, early and late multi-sensor fusion, and the state-of-the-art method, EcoFusion \cite{malawade2022ecofusion}.
To measure the object detection performance of each model, we use the object detection loss function from \cite{girshick2015fast}, which combines bounding box loss with classification loss. 
The object detection metrics we present are for a Faster R-CNN variant of our model trained using the same hyperparameters as \cite{malawade2022ecofusion} for fairer comparison with EcoFusion \cite{malawade2022ecofusion}. However, we verified experimentally that the SSD-based model achieves 50\% lower average loss and consumes 15\% less energy than the Faster R-CNN version. 
We used built-in functions in the host code and system commands to measure the end-to-end latency and power consumption of different configurations.

\subsection{Performance on FPGA}
We compare the object detection performance and energy consumption of different fusion techniques in Table \ref{tab:fpga_results}. Across different gating and $\lambda_E$ configurations, CARMA achieves lower average energy usage and loss than almost every early fusion, late fusion, and single sensor model. 
The only exceptions were the camera-only configurations, which had higher losses than our method but lower energy usage due to the efficiency of the camera sensors. 
Notably, with an equivalent model loss, CARMA ($\lambda_E=0, deep$) achieves a \textbf{41.3\%} reduction in energy compared to EcoFusion ($\lambda_E=0, attn$). With a higher $\lambda_E=0.01$ for both models, CARMA achieves \textbf{73.7\%} lower energy usage with only a 3.2\% higher loss than EcoFusion. EcoFusion's inability to account for sensor energy or apply sensor and model clock gating leads to higher average energy consumption, putting it on par with high-energy early fusion and late fusion variants. 
CARMA also exhibits faster speeds, achieving \textbf{6\%-33\%} speed-up compared to EcoFusion, with lower model latencies for higher $\lambda_E$ values. The results highlight trade-offs among sensing modalities, with radar branches consuming more energy but providing reliability in camera failure contexts, as supported by lower loss in the late fusion model.

\begin{table}[ht]
\footnotesize
    \centering
\begin{tabular}{p{30pt} p{70pt} p{30pt} p{30pt} p{30pt}}\hline
\textbf{Fusion Type} & \textbf{Configuration} & \textbf{Avg. Loss} & \textbf{Energy (J)} & \textbf{Latency (ms)}\\\hline
\multirow{3}{30pt}{None} & Radar ($R$) & 2.858 & 6.73 & 14.2\\
 & LiDAR ($L$) & 4.682 & 3.73 & 14.2\\
 & Camera ($C$)  & 1.680 & 1.81 & 14.2\\\hline
\multirow{3}{30pt}{Early} & $R + L$ & 2.784 & 9.16 & 17.1\\
 & $C_L$ + $C_R$ & 1.203 & 2.31 & 17.1\\
 & $L + C_L + C_R$ & 3.476 & 3.73 & 19.7\\\hline
Late & $R + L + C_L + C_R$ & 0.967 & 10.48 & 42.6\\\hline
\multirow{3}{30pt}{EcoFusion \cite{malawade2022ecofusion}} & $\lambda_E=0, attn$ & 0.915 & 10.41 & 54.0\\
 & $\lambda_E=0.01, attn$ & 0.924 & 10.36 & 48.0\\
 & $\lambda_E=0.1, attn$ & 1.147 & 10.18 & 27.7\\\hline
\multirow{6}{30pt}{\textbf{CARMA (Ours)}} &  $\lambda_E=0, attn$ & 0.915 & 7.35 & 51.9\\
 &  $\lambda_E=0, deep$ & 0.915 & 6.12 & 51.2\\
 &  $\lambda_E=0.0001, attn$ & 0.920 & 6.68 & 50.2\\
 &  $\lambda_E=0.001, deep$ & 0.944 & 3.31 & 42.6\\
 &  $\lambda_E=0.001, attn$ & \textbf{0.959} & \textbf{3.23} & \textbf{38.5}\\
 &  $\lambda_E=0.01, deep$ & \textbf{0.954} & \textbf{2.73} & \textbf{36.1}\\
\hline
\end{tabular}
    \caption{Performance and energy comparison between different fusion methods ($T_c$ = 30 samples)}
    \label{tab:fpga_results}
    \vspace{-1.5em}
\end{table}

\begin{figure}[ht]
    \centering
    \includegraphics[width=0.86\linewidth, clip]{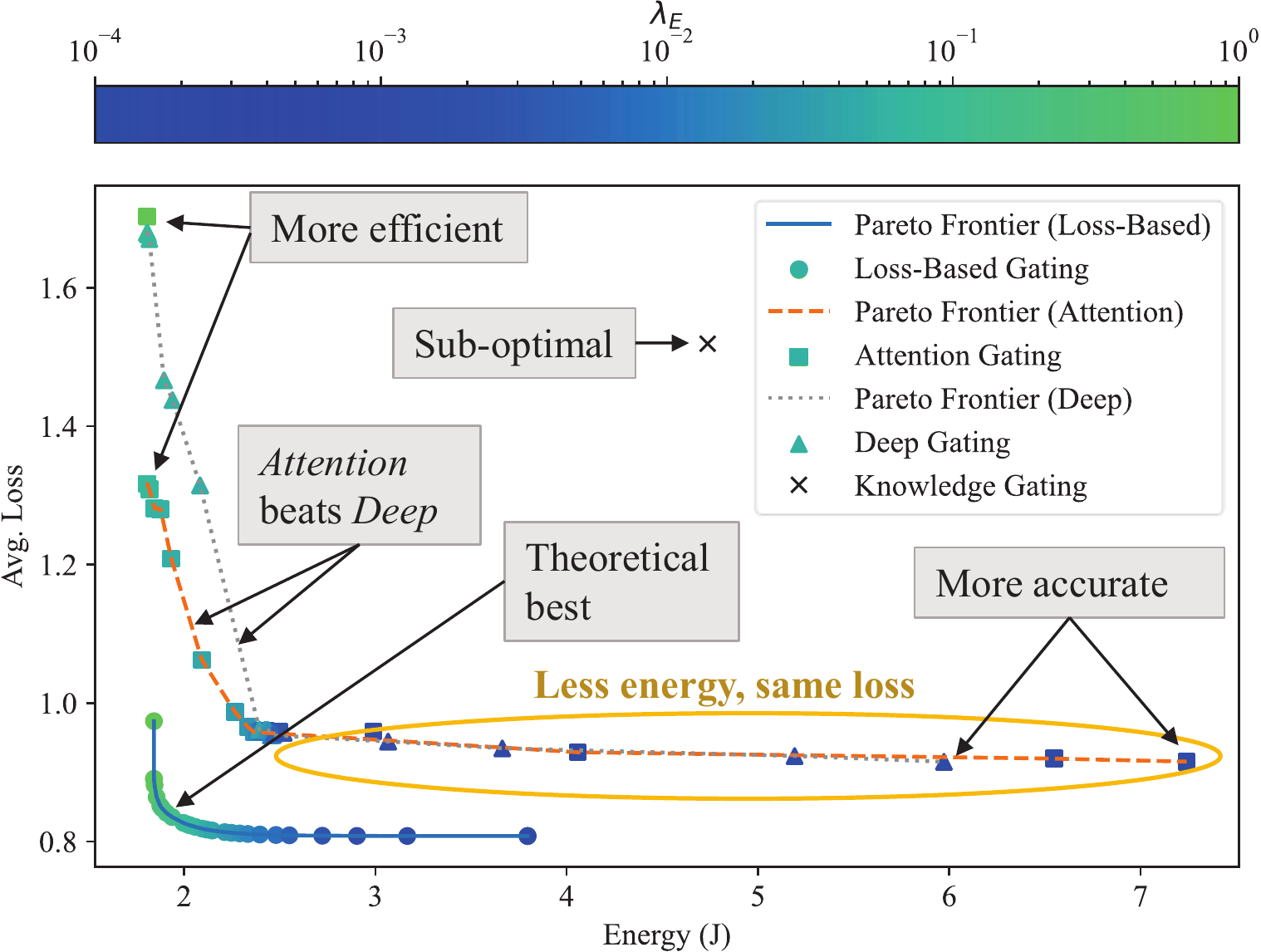}
    \caption{System-wide energy consumption vs. object detection loss of different gate modules for varying values of $\lambda_E$.}
    \label{fig:optim_result}
    \vspace{-3mm}
\end{figure}


\begin{figure}[!ht]
        \centering
        \begin{subfigure}{\linewidth}
            \centering
            \includegraphics[width=0.85\linewidth, clip]{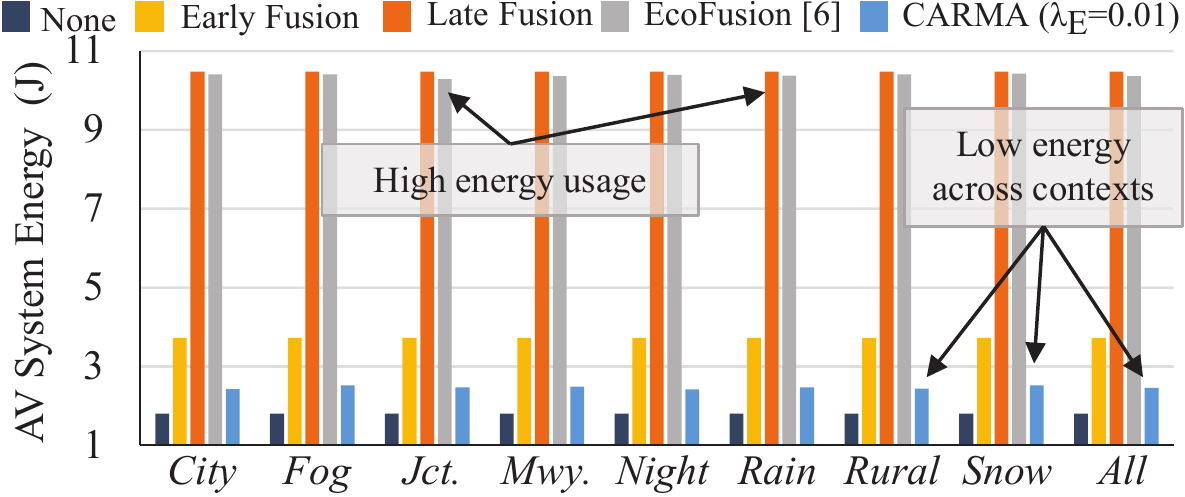}
        \end{subfigure}
        \hfill
        \begin{subfigure}{\linewidth}
            \centering
            \includegraphics[width=0.85\linewidth, clip]{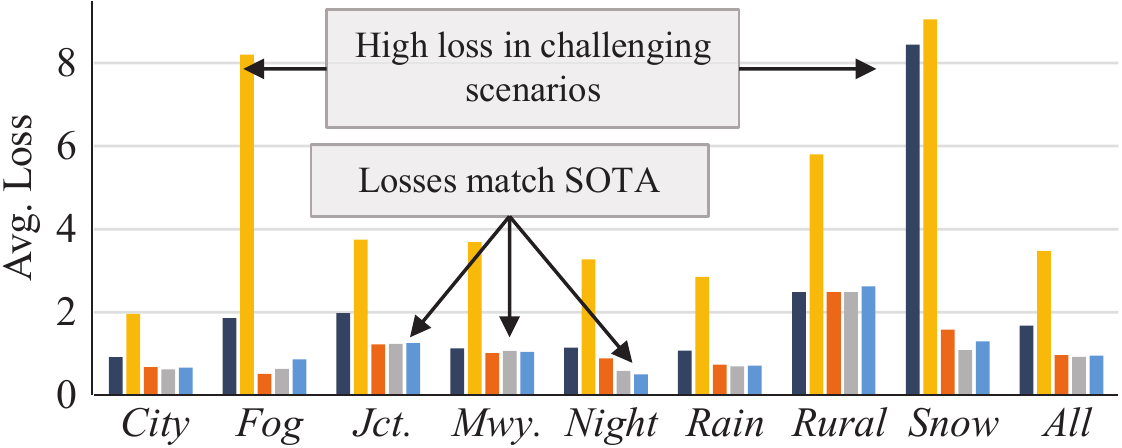}
            \begin{minipage}{.1cm}
            \vfill
            \end{minipage}
        \end{subfigure} 
        \caption{Scenario-specific energy usage and object detection loss for: No Fusion ($C_R$), Early Fusion ($L + C_L + C_R$), Late Fusion ($R+L+C_L+C_R$), EcoFusion ($\lambda_E=0, attn$), and CARMA ($\lambda_E=0.01, attn$).}\vspace{-1mm} 
        \label{fig:energy_loss_plot}
        \vspace{-4mm}
\end{figure}


Fig. \ref{fig:optim_result} illustrates the trade-off between system-wide energy consumption and model performance for each gate module at different values of  $\lambda_E$. 
Both \textit{deep} and \textit{attn} gates present a clear trade-off between performance and energy efficiency as $\lambda_E$ increases. However, the large flat region along the right side of both Pareto frontiers illustrates how system-wide energy can be reduced significantly with a minimal performance impact. 
The results for \textit{loss-based} gating indicate the performance of an optimal gate module and serve as a theoretical upper bound, since it uses the posteriori ground truth loss to select branch. The \textit{knowledge} gate is ineffective in minimizing either objective.
Overall, the \textit{deep} and \textit{attn} gate reduce energy consumption by over \textbf{55\%} while maintaining an average loss within 5\% of the $\lambda_E=0$ models.

\subsection{Scenario-Specific Performance}
Fig. \ref{fig:energy_loss_plot} shows how different driving scenarios affect the energy consumption and performance of different fusion methods. The results show that CARMA can reduce energy consumption below that of early fusion, late fusion, and EcoFusion across all scenarios. 
Interestingly, our model minimizes energy consumption in the \textit{Snow} scenario by selecting camera branches only throughout the context ($C_L$, $C_R$, and $C_L+C_R$).
Early fusion is especially weak in the \textit{Fog}, \textit{Rural}, and \textit{Snow} contexts, likely due to its susceptibility to sensor noise. Late fusion, EcoFusion, and CARMA are robust across all scenarios, with \textit{Rural} being the most challenging.

\section{Conclusion}
In this work, we proposed a context-aware sensor fusion approach that uses context to reconfigure the perception model on an FPGA at runtime dynamically. 
CARMA is capable of switching model computation paths with negligible latency while intermittent context identification, system-wide energy-performance optimization, and sensor clock gating maximize energy savings without compromising performance. Overall, CARMA achieves up to $1.3 \times$ speedup and reduces energy consumption by over 73\% over leading static and dynamic sensor fusion techniques across complex driving contexts. 

\section*{Acknowledgment}
This work was partially supported by the NSF under award CCF-2140154 and hardware donations from AMD-Xilinx University Program.

{\tiny
\bibliographystyle{IEEEtran}
\bibliography{IEEEabrv, bibliography}}

\end{document}